\documentclass[10pt,twocolumn,letterpaper]{article}

\usepackage[pagenumbers]{cvpr} % To force page numbers, e.g. for an arXiv version

\usepackage{algorithm}
\usepackage{algpseudocode}

\definecolor{cvprblue}{rgb}{0.21,0.49,0.74}
\usepackage[pagebackref,breaklinks,colorlinks,allcolors=cvprblue]{hyperref}

\def\paperID{*****} % *** Enter the Paper ID here
\def\confName{CVPR}
\def\confYear{2026}

\title{3$^{\mathrm{rd}}$ Place at CVPR 2026 CASTLE Challenge: Agentic Multi-View Long-Context Video Understanding via Hierarchical Knowledge Graph Retrieval}

\author{
Raghad Albusayes\thanks{Equal contribution.} \qquad Munirah Alyahya\footnotemark[1] \\
TAHAKOM \\
Riyadh, Saudi Arabia \\
{\tt\small ralbossaius@tahakom.com \qquad malyahya@tahakom.com}
}

\begin{document}
\maketitle
\begin{abstract}
This paper presents our winning methodology for the CASTLE 2026 Challenge at the CVPR 2026 EgoVis Workshop, where our team secured third place globally. The challenge tasks participants with answering highly complex visual, spatiotemporal, and verbal questions, including visual counting, action localization, multi-view tracking and speaker temporal reasoning, within massive, multimodal video streams. The underlying dataset consists of over 600 hours synchronized footage captured by 15 ego and exo camera sources. To tackle the extreme scale and long-context demands of this environment, we introduce a training-free agentic framework optimized for long-form video understanding. Our framework introduces two core architectural components: i) a Video Knowledge Graph that maps static and dynamic entities, their temporal relationships, and intersecting events to enable multi-hop relational reasoning, and ii) an adaptive agentic workflow that resolves complex queries through a hierarchical retrieval and indexing. Empirical results demonstrate that our framework achieves high zero-shot reasoning accuracy on long-context multi-view streams. Our code will be released at \url{https://github.com/RaghadKhaled/CASTLE-Challenge-Framework}.
\end{abstract}    
\section{Introduction}
\label{sec:intro}

Long-form videos have become a dominant medium across domains such as movies, CCTV monitoring, and egocentric AI assistants. However, long video understanding remains highly challenging, as it requires the capability to reason over spatiotemporal details within long contexts. Existing approaches for long video understanding can be broadly categorized into three paradigms. The first is single-pass Multi-modal LLMs (MLLMs)-based methods~\cite{ref_1, ref_2, ref_3}, which directly process video frames and rely on key frame selection, extended context windows~\cite{ref_4}, or token compression~\cite{ref_5} to fit long videos within model constraints. The second is Retrieval-Augmented Generation (RAG)-based methods~\cite{ref_6, ref_7, ref_8}, which represent video memory as either chunk-based or graph-based structures and retrieve relevant information for reasoning. The third is agentic-based methods~\cite{ref_9, ref_10, ref_11, ref_12}, which decompose video understanding into iterative perception, reasoning, and action processes over a video to progressively gather information and produce answers. Building upon prior work~\cite{ref_12}, we further enhance both the graph construction and agentic framework, and propose a multimodal, multi-camera framework for long video QA. Our approach represents long videos as structured temporally annotated graphs that model static and dynamic entities together with their relationships, enabling complex multi-hop reasoning across long temporal contexts. In addition, we design a hierarchical agentic retrieval module for answering queries through multi-level memory retrieval. Our framework achieved third place in the CVPR 2026 CASTLE Challenge, establishing a new state-of-the-art in long-video understanding.
%-------------------------------------------------------------------------

% \subsection{Miscellaneous}

% However, use it only when there are three or more authors.
% Thus, the following is correct:
%    ``Frobnication has been trendy lately.
%    It was introduced by Alpher~\cite{Alpher02}, and subsequently developed by

\section{Methodology}
\label{sec:methodology}

Our proposed framework is training-free and builds upon the approach introduced in~\cite{ref_12}. The methodology is structured as follows: Section~\ref{subsec:Problem_Formulation} formalizes the task definition. Section~\ref{subsec:Memory_Construction} illustrates the multimodal graph construction stage. Finally, Section~\ref{subsec:Agentic_Framework} details the agentic framework.

% \begin{figure*}[t]
%     \centering
%     \fbox{\rule{0pt}{3in}\rule{0.95\textwidth}{0pt}}
%     \caption{Your figure caption.}
%     \label{fig:main}
% \end{figure*}

\begin{figure*}[t]
    \centering
    \includegraphics[width=\textwidth]{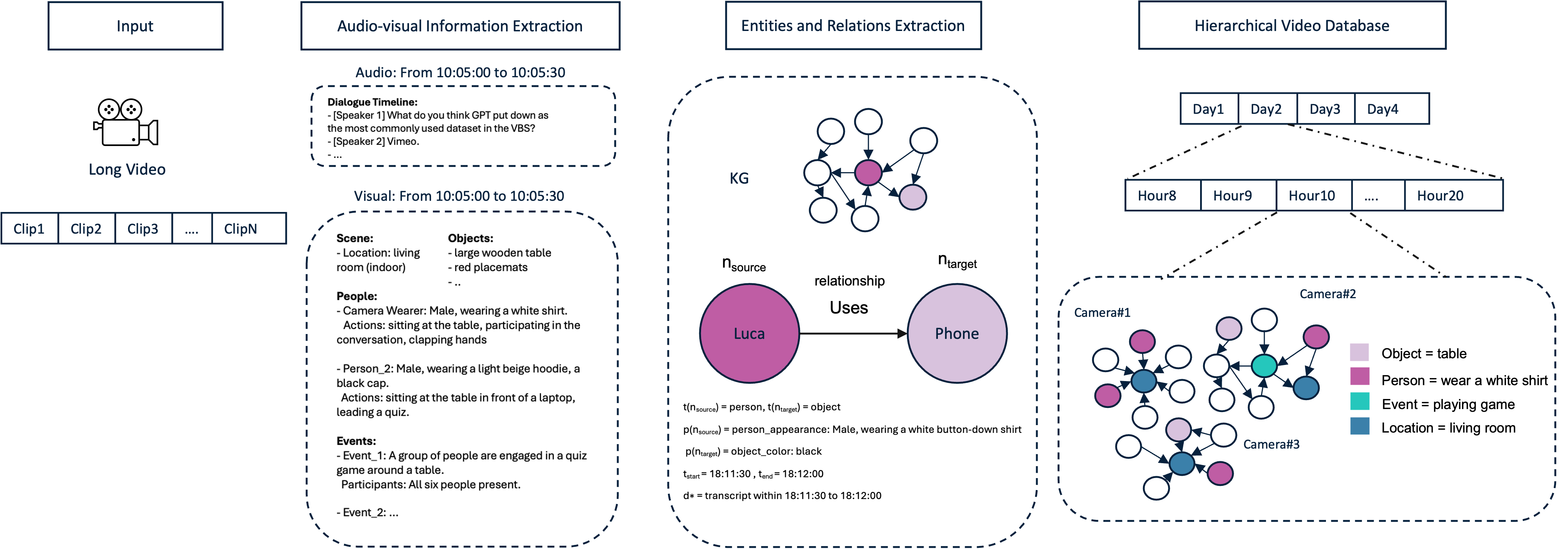}
    \caption{Overview of memory construction stage.}
    \label{fig:main}
\end{figure*}

%-------------------------------------------------------------------------
\subsection{Problem Formulation}
\label{subsec:Problem_Formulation}
Given a collection of one-hour video recordings with audio $V = \{v_1, v_2, \ldots, v_N\}$ captured over 4 days (12 hours per day) via 10 first-person and 5 third-person perspective cameras, our goal is to solve a long-form video question-answering task. During inference, the task is defined as follows: given a natural language query \textit{Q} and a set of candidate answers, the objective is to select the correct answer \textit{A}. In contrast to standard VQA tasks that are restricted to short, isolated video context, this formulation requires searching across more than 600 hours of long-form videos to localize the specific visual, temporal, or verbal content required for reasoning. We address this problem using a training-free, agentic framework.

\subsection{Memory Construction}
\label{subsec:Memory_Construction}
We introduce a multimodal memory construction approach that represents the videos in \textit{V} as a graph of static and dynamic entities, as illustrated in Figure~\ref{fig:main}.

{\bf Unified audio-visual information Extraction Stage.} Our analysis indicates that the egocentric streams captured by multi-ego camera setups sufficiently encapsulate scene-level information, allowing our framework to omit exocentric video inputs entirely. Formally, each video is uniformly divided into a set of 30-second clips, $S = \{s_1, s_2, \ldots, s_K\}$. To process these clips, we employ an omni-modal information extraction function $f_{\text{omni}}$, which jointly processes both visual and audio modalities in a single forward pass:

\begin{equation}
c_k = f_{\text{omni}}(s_k)
\end{equation}

where $c_k$ represents the multi-modal extracted description, including visual (scenes, objects, individuals, locations), temporal (actions, events), and verbal (transcripts).  The overall aggregated set of all clip-level captions across the long-form video sets is formalized as  $C = \{H_n\}_{n=1}^{N}$, where each hour subset is defined as  $H_n = \{C_{n,k}\}_{k=1}^{K}$.

{\bf Knowledge Graph Construction Stage.} Because isolated clip-level descriptions fail to capture the continuity of static and dynamic entities across temporal intervals, we construct a Knowledge Graph (KG) from the clip-level captions \textit{C}. The KG explicitly maps static and dynamic entities alongside their visual (attribute-based, e.g., located in), temporal (event-based, e.g., interacts with), and verbal relationships (e.g. mentions). By structuring this multimodal information into a unified KG, the framework enables complex, multi-hop relational reasoning across long-form video contexts.

\begin{algorithm*}[t]
\caption{Agentic Framework}
\label{alg:egagent}
\begin{algorithmic}[1]

\Require User query $Q$, KG, Videos $V$
\Ensure Answer $A$ to $Q$

\State Initialize global working memory $\mathcal{M}_g \leftarrow \emptyset$ and local working memory $\mathcal{M}_l \leftarrow \emptyset$

\State \textbf{Step 1: Query Decomposition}

% \State $\text{SubtaskList} \leftarrow \texttt{PlanningAgent.decompose}(Q)$ \hfill \{ $\text{SubtaskList} = \{(S_1,q_1), (S_2,q_2), \ldots, (S_N,q_N)\}$ \}
\State $\text{SubtaskList} \leftarrow \texttt{PlanningAgent.decompose}(Q)$ \quad \{ $\text{SubtaskList} = \{(S_1,q_1), (S_2,q_2), \ldots, (S_N,q_N)\}$ \}

\For{each $(S,q)$ in SubtaskList}

    \State \textbf{Step 2a: Retrieve relevant data for the subtask}

    \State $\text{sql\_q} \leftarrow \texttt{GraphSearch.generate\_sql\_queries}(q)$ 
    
    \State $\text{RetrievedData} \leftarrow \texttt{GraphSearch.execute\_sql\_queries}(sql\_q)$

    % \State $\text{RetrievedData} \leftarrow \texttt{GraphSearch.executesql_queries}(sqlq)$
    
    % \State $\text{RetrievedData} \leftarrow T(q)$
    
    % \Statex \hspace{1em} {Visual: hybrid semantic/attribute search; Audio: transcript search; Entity Graph: SQL queries}

    \State \textbf{Step 2b: Update working memories}

    \State $\mathcal{M}_g \leftarrow \mathcal{M}_g \cup \{\text{RetrievedData}\}$

    \State $\mathcal{M}_l \leftarrow \{\text{RetrievedData}\}$

    \State \textbf{Step 3: Decide whether we need ClipSearch for the subtask}

    % \State $\text{Decision} \leftarrow \texttt{Reflector.assessment}(\text{q, Q, \mathcal{M}_l)$

    \State $\text{Decision} \leftarrow \texttt{Reflector.assessment}(q, Q, \mathcal{M}_l)$

    \If{$\text{Decision} = \text{``yes''}$}
        \State \textbf{Step 3a: Retrieve clips data for the subtask}

        \State $\text{ClipsData} \leftarrow \texttt{ClipSearch.select\_clips}(q, \mathcal{M}_l)$

        \State $\text{RetrievedClips} \leftarrow \texttt{ClipSearch.retrieve}(ClipsData)$

        \State \textbf{Step 3b: Analyze retrieved clips for relevance and evidence}
        
        \State $\text{Visual\_Analysis} \leftarrow \texttt{ClipAnalyze.analyze}(RetrievedClips, q)$

        \State \textbf{Step 3c: Update working memories}

        \State $\mathcal{M}_g \leftarrow \mathcal{M}_g \cup \{\text{Visual\_Analysis}\}$        
        % \State \textit{\# Your next steps go here, automatically indented}
    \EndIf

    % \State $\text{Analysis} \leftarrow \texttt{AnalyzerTool.analyze}(\text{RetrievedData}, S)$
    
    % \Statex \hspace{1em} {LLM-based reasoning, evidence extraction, filtering}

    % \Statex \textbf{Step 2c: Update working memory}

    % \State $\mathcal{M} \leftarrow \mathcal{M} \cup \{\text{Analysis}\}$

\EndFor

\State \textbf{Step4 : Answer Generation}

\State $A \leftarrow \texttt{VQAAgent.answer}(Q,\mathcal{M}_g)$

% \Statex \hspace{1em} {VQAAgent uses accumulated cross-modal evidence in $\mathcal{M}$ to answer $Q$}

\State \Return $A$

\end{algorithmic}
\end{algorithm*}

\textit{Graph Design.}
We construct a directed Knowledge Graph $G = (V, E)$, where:

\begin{itemize}
    \item $Nodes (N)$: a set of entities $\{n_1, n_2, \ldots, n_K\}$  and their types $T \subseteq \mathbb{N} \times \mathbb{N}$.
    \item $Edges (E)$: a set of relationships $R \subseteq \mathbb{N} \times \mathbb{N}$, along with a temporal interval $[t_{\text{start}}, t_{\text{end}}]$ represent the timestamp of the relationship.
\end{itemize}

% •	$Nodes (N)$: a set of entities $\{n_1, n_2, \ldots, n_K\}$  and their types T  N × N.
% •	$Edges (E)$: a set of relationships R  N × N, along with a temporal interval $[t_start, t_end]$ represent the timestamp of the relationship.

Where each node $n \in \mathbb{N}$ has type $t \in \mathbb{T}$. The set of entity types is:
\begin{equation}
\begin{aligned}
T = \{ & \text{Person}, \text{Location}, \text{Object}, \text{Food}, \text{Activity} \\
       & \text{Event}, \text{Action} \}
\end{aligned}
\end{equation}

And each edge $e \in \mathbb{E}$ has a relationship type $r \in \mathbb{R}$. The set of relationship types is:
\begin{equation}
\begin{aligned}
T = \{ & \text{talks-to}, \text{interacts-with}, \text{mentions}, \text{uses}, \\
       & \text{located-in}, \text{occurs-at}, \text{doing}, \text{is} \}
\end{aligned}
\end{equation}

Each relationship of an edge could have one or more temporal intervals as a property. It represented when this relationship happened in a video. The entity nodes are enriched with type-specific properties \textit{P}. The designated property types are defined as follows:
\begin{equation}
\begin{aligned}
P = \{ & \text{location-type (for Location)}, \text{object-color (for Object)}, \\
    & \text{person-appearance (for Person)} \}
\end{aligned}
\end{equation}

The graph edges encapsulate a comprehensive taxonomy of multimodal interactions. Specifically, these relationship types characterize granular dependencies across the clips within a video, including verbal communication (either direct human-to-human dialogue via talks-to relation, or general, non-directed utterances via mentions relation), spatial grounding (person–location, object–location), physical manipulation (person–object/food), and behavioral interactions (person–action/event/activity).

\textit{Visual-Verbal Entities and Relations Extraction.}
For each video $H_n$ (one hour duration as an example), we construct its KG, $G = (V, E)$ , using Langchain’s LLMGraphTransformer~\cite{ref_13}. Given the set of captions for every clip within this video, $H_n = \{C_{n,k}\}_{k=1}^{K}$, we formulate the context as a set of 30-second clip captions with their corresponding timestamps. Incorporating timestamps with captions enables each edge in the graph to associate relationships with the relevant temporal intervals. Each caption $c_k$ is divided into verbal context and visual/temporal context to extract relationships separately, allowing the LLMGraphTransformer~\cite{ref_13} to focus on each contextual domain independently.
We edit the implementation of LLMGraphTransformer~\cite{ref_13}  to support the extraction of additional properties to the nodes and edges (temporal annotations of the relationships) at the same stage of the graph extraction process. We also experimented with the temporal annotation method proposed in~\cite{ref_12}, implementing it as a standalone stage after graph construction.
The final output graph of a video is $G_n = (V_n, E_n)$, where each edge is formatted as a tuple of:
\begin{equation}
\begin{aligned}
( & n_\text{source}, t(n_\text{source}), n_\text{target}, t(n_\text{target}), p(n_\text{source}), p(n_\text{target}), \\
    &  t_\text{start}, t_\text{end}, \text{$d^*$} )
\end{aligned}
\end{equation}

where $n_\text{source}$ and $n_\text{target}$ are the source and target nodes, and  $d^*$ denotes any information within the temporal interval of the edge. Additional information can be incorporated within this interval when annotations are available (e.g. transcripts). 
The final graph across all videos is stored in a single SQLite3 memory database, where each row corresponds to one tuple.

\textit{Hierarchical Spatial-Temporal Retrieval Granularity }
To enable multi-level hierarchical retrieval, each edge is annotated with a temporal interval and metadata (Day, Hour, Camera name) to support direct access to visual evidence. This forms a multi-level temporal memory that aggregates multiple graphs within the same camera, hour, and day. For each entity type $t \in \mathbb{T}$, a super node represents a global set of unique prototype entities that group all nodes belonging to the same entity type. This entity type of the nodes supports hierarchical retrieval using complex queries across temporal dimensions (e.g., day, hour, events, actions) and spatial dimensions (e.g., location, object). Associating each edge with a camera name also enables multi-camera reasoning when the same event, action, or dialogue occurs at the same location across different cameras. The edge representation in the database is defined as follows:
\begin{equation}
\begin{aligned}
( & \text{day}, \text{hour}, \text{camera name}, n_\text{source}, t(n_\text{source}), n_\text{target}, t(n_\text{target}), \\
    & p(n_\text{source}), p(n_\text{target}), t_\text{start}, t_\text{end}, \text{$d^*$} )
\end{aligned}
\end{equation}

\subsection{Agentic Framework}
\label{subsec:Agentic_Framework}
Our Agentic framework is present in Algorithm~\ref{alg:egagent}. Given a query \textit{Q}, KG, and video list \textit{V}, the agent framework divides the \textit{Q} into \textit{N} sub-tasks and collects relevant evidence from the KG or row videos in an iterative process to generate the final answer. The framework consists of 4 main components:

{\bf PlannerAgent:} The Planner agent (Algorithm~\ref{alg:egagent}, Line 2-3) first decomposes the user query \textit{Q} into \textit{N} natural language steps $\{1_1, 1_2, \ldots, q_N\}$. Each step $qi$ represents a distinct video or speech QA task, encompassing temporally constrained localization (e.g., by hour or day), action recognition, object attribution, spoken phrase detection, and transcript keyword searching.

{\bf GraphRetrival:} Each plan step $q_i$ is passed to the GraphRetrival Agent to retrieve relevant information from the KG  (Algorithm~\ref{alg:egagent}, Line 5 to 7) and save it to the working memories (Algorithm~\ref{alg:egagent}, Line 8 to 10). The GraphRetrieval Agent initially constructs SQL queries tailored to $q_i$ employing a 'strict-to-relaxed' generation strategy. Under this hierarchical approach, the agent first generates a stringent SQL query matching exact database fields. If no records are returned, it gradually relaxes the SQL query constraints to ensure a broader retrieval of relevant rows.

{\bf ClipRetrieval and ClipAnalyze:} The Reflector Agent (Algorithm~\ref{alg:egagent}, Lines 11-12) determines if the current planning step $q_i$ requires clip retrieval by evaluating the query \textit{Q} and memory context $M_l$ for missing visual data. If required, the ClipRetrieval Agent (Algorithm~\ref{alg:egagent}, Line 14-20) utilizes the temporal fields of relevant rows in $M_l$ to extract localized clips. Consequently, the KG serves as a structured temporal index to guide video boundary detection. The ClipAnalyze Agent then evaluates these clips and uses $q_i$ as a question to deliver an answer supported by visual evidence.

{\bf VQA Agent:} The VQA Agent produces the final answer \textit{A} based on original query \textit{Q} and the accumulated evidence in $M_g$ (Algorithm~\ref{alg:egagent}, Line 23-24). The VQA agent implicitly analyzes the working memory context and outputs its detailed reasoning process to reach out the correct answer.

\section{Experiments}
\label{sec:experiments}

We evaluate the performance of our proposed framework against a baseline on the CASTLE benchmark~\cite{ref_14}, which serves as the official benchmark used in the CASTLE Challenge at the CVPR 2026 EgoVis Workshop. We first provide an overview about the benchmark in Section~\ref{subsec:Evaluation_Benchmark}, then we illustrate the implementation details in Section ~\ref{subsec:Implementation_Details}, followed by a performance analysis against the baseline as well as a discussion about analyzing the impact of key factors in the proposed approach in Section ~\ref{subsec:Results_and_Discussion}.

%-------------------------------------------------------------------------
\subsection{Evaluation Benchmark}
\label{subsec:Evaluation_Benchmark}
The CASTLE benchmark~\cite{ref_14} is a large-scale, multimodal dataset designed for advancing research in lifelogging, human activity recognition, and multimodal retrieval. The version of the benchmark adopted for the challenge at the CVPR 2026 EgoVis Workshop consists of 185 closed-form long-context question-answering samples with 600 hours of synchronized videos supplemented by audio across four days by 15 egocentric (first-person) and exocentric (third-person) camera sources. The questions are formulated in the past tense and are designed to assess a comprehensive set of multimodal capabilities spanning visual, temporal, and verbal domains. Specifically, they evaluate skills such as counting, optical character recognition (OCR), spatial relationship reasoning, object attribute recognition, object and action localization, activity recognition, multi-view tracking, person identification, speaker identification, speaker temporal reasoning, and multi-turn dialogue reasoning.

\subsection{Implementation Details}
\label{subsec:Implementation_Details}
Our proposed method was implemented using the open-source LangGraph~\cite{ref_13} library. For all pipeline stages, we utilized the Gemini 2.5 Pro model configured with default settings, a temperature of 0, and a maximum of 3 retries. The maximum number of planning steps was set to 5. For the unified Audio-Visual captioning stage from video clips, the original 50 fps input videos were sampled to 1 fps while maintaining their original resolution of 3840 × 2160. The audio of the video was preserved during this multimodal information extraction stage. However, in the Clip Analysis agent, the audio was removed from the video to focus exclusively on the analysis of the visual content.
For the Knowledge Graph construction stage, we use Langchain’s LLMGraphTransformer~\cite{ref_13} to extract the nodes, node properties, edge relationships, and edge properties from our generated unified audio-visual captions.

\begin{table}
  \caption{Comparison with the baseline. $^*$We eliminate the audio and visual databases with their analysis agents. The two-step process of visual captioning followed by visual transcript fusion is replaced with unified audio-visual captioning as a single step.}
  \centering
  \begin{tabular}{@{}lc@{}}
    \toprule
    Method & Accuracy (\%) \\
    \midrule
    Baseline~\cite{ref_12}$^*$ & 43 \\
    \textbf{Ours} & \textbf{55} \\
    \bottomrule
  \end{tabular}
\label{tab:table_1}
\end{table}

% \begin{table}[t]
% \caption{Comparison with the baseline. $^*$Note: Evaluated using the zero-shot protocol variant.}
% \leftline{\small $^*$Note: Evaluated using the zero-shot protocol variant.}
% \vspace{1mm}
% \centering
% \begin{tabular}{lc}
% \hline
% Method & Accuracy \\
% \hline
% Baseline~\cite{ref_12}$^*$ & 43 \\
% Ours & \textbf{55} \\
% \hline
% \end{tabular}
% \label{tab:table_1}
% \end{table}

\begin{table}
  \caption{Impact of KG DB retrieval content on working memory.}
  \centering
  \begin{tabular}{@{}lc@{}}
    \toprule
    Method & Accuracy (\%) \\
    \midrule
    Our (top 5 rows) & 49 \\
    \textbf{Our (top 50 rows)} & \textbf{52} \\
    Our (top 250 rows) & 51 \\
    \bottomrule
  \end{tabular}
\label{tab:table_3}
\end{table}

\begin{table}
  \caption{Effect of including "Mention" relationships in the KG database.}
  \centering
  \begin{tabular}{@{}lc@{}}
    \toprule
    Method & Accuracy (\%) \\
    \midrule
    \textbf{Our (with Mention)} & \textbf{51} \\
    Our (without Mention) & 44 \\
    \bottomrule
  \end{tabular}
\label{tab:table_4}
\end{table}

\subsection{Results and Discussion}
\label{subsec:Results_and_Discussion}
Table~\ref{tab:table_1} presents a comparison between the baseline~\cite{ref_12} and our proposed approach. Following the enhancements made to our graph construction and agentic framework, our proposed approach achieved an accuracy of 55\%, marking a +12\% improvement over the baseline.  

% \subsection{Ablation Study}
% \label{subsec:Ablation_Study}
In order to evaluate the individual performance contributions of key components within our framework, we conducted a systematic ablation study focusing on three primary factors: the number of retrieved KG database rows retained in the working memory, the inclusion versus exclusion of database rows containing "Mention" relationships, and the integration of the ClipSearch and ClipAnalysis module into the framework.

Table~\ref{tab:table_3} illustrates the performance impact of varying the number of retrieved KG database rows saved in the working memory. Because these rows contain dense transcript data, accumulating rows across multiple planning steps can cause context window overhead, potentially exhausting the input token limit for VQA Agent. To mitigate this token overhead, we evaluate a filtering strategy that preserves only the top-\textit{N} retrieved rows. Increasing \textit{N} from 5 to 50 improves accuracy by 3\%, demonstrating that a larger context provides critical details or it can contain the relevant answer to the plan step. However, scaling beyond this threshold degraded performance, as trailing rows often introduce irrelevant noise that increases model hallucination during final answer generation.

Table~\ref{tab:table_4} evaluates the performance impact of including versus excluding the database with the "Mention" relationship. Our initial hypothesis assumes that these rows could be excluded because the corresponding transcript column already encapsulates the spoken dialogue, allowing the GraphRetrival agent to search and rely on the textual content of the transcript column. However, the empirical results demonstrate that "Mention" relations provide critical structural data to search for specific words and specify the precise speaker identity (if it is the camera wearer) and localized temporal intervals. In contrast, the transcript entries aggregate dialogue across multiple speakers, with highly variable intervals depending on the relationship type and its source and target node types. Furthermore, because the rows in the database depend on the specific interval defined by the relationship, excluding "Mention" relationship risks data loss, leaving clips of the dialogue unrepresented within the database.

Table~\ref{tab:table_5} evaluates the performance impact of integrating the ClipSearch and ClipAnalyze modules into the agentic framework. We hypothesized that certain query categories, such as those requiring visual counting, OCR, or spatial reasoning, cannot be answered using textual metadata within the KG database alone. Instead, the structured information in the KG database serves as a temporal index, guiding to the precise video and clip boundaries required for localized retrieval. Adding ClipSearch and ClipAnalyze modules with excluding non-visual relationships (”Mention”) to handle these visual queries yields a performance boost of 4\%. 

These findings highlight the incremental contribution of the mentioned factors to the overall effectiveness of our approach.

Finally, Table~\ref{tab:table_2} summarizes the public leaderboard with the accuracy scores of the top five teams in the CASTLE Challenge at the CVPR 2026 EgoVis Workshop.

\begin{table}
  \caption{Performance impact of the ClipSearch module.}
  \centering
  \begin{tabular}{@{}lc@{}}
    \toprule
    Method & Accuracy (\%) \\
    \midrule
    Our (without ClipSearch) & 51 \\
    \textbf{Our (With ClipSearch + without Mention)} & \textbf{55} \\
    \bottomrule
  \end{tabular}
\label{tab:table_5}
\end{table}

\begin{table}
  \caption{Comparison with the Leaderboard.}
  \centering
  \begin{tabular}{lcc}
    \toprule
    Method & Rank & Accuracy (\%) \\
    \midrule
    WDL & 1 & 58 \\
    ilearn\_zhy & 2 & 57 \\
    \textbf{TAHAKOM} & \textbf{3} & \textbf{55} \\
    SB-CuriousAI & 4 & 50 \\
    Vitality & 5 & 21 \\
    \bottomrule
  \end{tabular}
\label{tab:table_2}
\end{table}

\section{Conclusion}

In this report, we introduce a multi-modal, multi-camera agentic framework for the long video QA task. Our framework represents the videos in Knowledge Graph with hierarchical retrieval granularity to enable multi-level, multi-hop reasoning for complex questions in the Castle Benchmark. Experimental results demonstrate that our approach achieves strong performance, securing third place in CVPR 2026 CASTLE Challenge and surpassing previous state-of-the-art approaches. Future work will focus on optimizing the memory design, exploring different open- and closed-source models, improving the agentic framework modules, and evaluating the domain-agnostic capability of the approach across other long video QA benchmarks.

{
    \small
    \bibliographystyle{ieeetr}
    \bibliography{main}
}


\end{document}